\documentclass[12pt]{colt2021}

\usepackage{graphicx}

\newtheorem{problem}{Problem}

\DeclareMathOperator*{\argmin}{argmin}

\DeclareMathOperator*{\cost}{cost}

\DeclareMathOperator*{\expect}{\mathbb{E}}

\title{Regret-optimal control in dynamic environments}

\coltauthor{Gautam Goel \Email{ggoel@caltech.edu}\\
 \addr California Insititute of Technology
\AND Babak Hassibi \Email{hassibi@caltech.edu}\\ \addr California Insititute of Technology}

\begin{document}

\maketitle


\begin{abstract}
    We consider control in linear time-varying dynamical systems from the perspective of regret minimization. Unlike most prior work in this area, we focus on the problem of designing an online controller which minimizes regret against the best dynamic sequence of control actions selected in hindsight (dynamic regret), instead of the best fixed controller in some specific class of controllers (policy regret). This formulation is attractive when the environment changes over time and no single controller achieves good performance over the entire time horizon. We derive the state-space structure of the regret-optimal controller via a novel reduction to $H_{\infty}$ control and present a tight data-dependent bound on its regret in terms of the energy of the disturbance. Our results easily extend to the model-predictive setting where the controller can anticipate future disturbances and to settings where the controller only affects the system dynamics after a fixed delay.  We present numerical experiments which show that our regret-optimal controller interpolates between the performance of the $H_2$-optimal and $H_{\infty}$-optimal controllers across stochastic and adversarial environments. 
\end{abstract}

\vspace{2mm}


\section{Introduction}
The central question in control theory is how to regulate the behavior of an evolving system with state $x$ that is perturbed by a  disturbance $w$ by dynamically adjusting a control action $u$. Traditionally, this question has been studied in two distinct settings: in the $H_2$ setting, we assume that the disturbance $w$ is generated by a stochastic process and seek to select the control $u$ so as to minimize the \textit{expected} control cost, whereas in the $H_{\infty}$ setting we assume the noise is selected adversarially and instead seek to minimize the \textit{worst-case} control cost.

Both $H_2$ and $H_{\infty}$ controllers suffer from an obvious drawback: they are designed with respect to a specific class of disturbances, and if the true disturbances fall outside of this class, the performance of the controller may be poor. Indeed, the loss in performance can be arbitrarily large if the disturbances are carefully chosen \cite{doyle1978guaranteed}.

This observation naturally motivates the design of \textit{adaptive} controllers, which dynamically adjust their control strategy as they sequentially observe the disturbances instead of blindly following a prescribed strategy. This problem has attracted much recent attention in the online learning community (e.g. \cite{agarwal2019online, hazan2020nonstochastic, foster2020logarithmic}), mostly from the perspective of regret minimization. In this framework, the online control policy $\pi$ is chosen so as to minimize  \textit{policy regret}: 
$$
\sup_{w} \left[ \cost(w, \pi(w)) - \inf_{\pi^* \in \Pi} \cost(w, \pi^*(w)) \right],
$$
where $\Pi$ is some class of policies and $\cost(w, u)$ is the control cost incurred by selecting the control action $u$ in response to the disturbance $w$. The comparator class $\Pi$ is often taken to be the class of state-feedback policies or the class of disturbance-action policies introduced in \cite{agarwal2019online}. The resulting controllers are adaptive in the sense that they seek to minimize cost irrespective of how the disturbances are generated.

In this paper, we take a somewhat different approach to the design of adaptive controllers. Instead of designing an online policy to minimize regret against the best policy selected in hindsight from some specific class, we instead focus on designing a policy which minimizes \textit{dynamic regret}. In other words, we focus on designing online controllers which compete against the optimal dynamic sequence of control actions selected in hindsight $$u^* = \argmin_{u} \cost(w, u). $$ We emphasize the key distinction between policy regret and dynamic regret: in policy regret, we compare the performance of the online policy to the single best fixed policy from a class of policies $\Pi$, whereas in dynamic regret we compare to a time-varying sequence of control actions, without reference to any specific class.

We believe that this dynamic regret formulation of online control is more attractive than the standard policy regret formulation for two fundamental reasons. Firstly, it is more general: instead of imposing \textit{a priori} some parametric structure on the controller we learn (e.g. state feedback policies, disturbance action policies, etc), which may or may not be appropriate for the given control task, we compete with the globally optimal dynamic sequence of control actions, with no artificial constraints. Secondly, and perhaps more importantly, the controllers we obtain are more robust to changes in the environment. Consider, for example, a scenario in which an online controller is trying to compete with the best state feedback controller selected in hindsight, and suppose the disturbances alternate between being generated by a stochastic process and being generated adversarially. When the disturbances are stochastic, an optimistic controller (such as the $H_2$ controller) will perform well; conversely, when the disturbances are adversarial, a more conservative, pessimistic controller (such as an $H_{\infty}$ controller) will perform well. No single state-feedback controller will perform well over the entire time horizon, and hence any online algorithm which tries to learn the best fixed controller will incur high cumulative cost. A controller which minimizes regret against the optimal dynamic sequence of control actions, however, is not constrained to converge to any fixed controller, and hence can potentially outperform standard regret-minimizing control algorithms  when the environment is dynamic.

Our approach to regret minimization in control is similar in outlook to a series of works in online learning (e.g. \cite{herbster1998tracking, hazan2009efficient, jadbabaie2015online,  goel2019beyond}), which seek to design online learning algorithms which compete with a dynamic sequence of actions instead of the best fixed action selected in hindsight; this formulation is natural when the reward-generating process encountered by the online algorithm varies over time. Unlike these prior works, we consider online learning in settings with \textit{dynamics}. This setting is considerably more challenging to analyze through the lens of regret, because the dynamics serve to couple costs across rounds; the actions selected by a learning algorithm in each round affect the costs incurred in all subsequent rounds, making counterfactual analysis difficult. 

It is well-known that dynamic regret can scale linearly in the number of rounds in the worst case (\cite{jadbabaie2015online}). For this reason, dynamic regret bounds are usually stated in a ``data-dependent" manner, where the regret is bounded in terms of the variation or energy of the input sequence. For example, \cite{besbes2015non, wei2018more} and \cite{bubeck2019improved} give dynamic regret bounds for bandit optimization in terms of the pathlength $\sum_{t=}^{T} \| \ell_t - \ell_{t-1}\|$, which measures the variation in the sequence of losses $\ell_1, \ldots \ell_T$; \cite{goel2019online} also give a dynamic regret bound in terms of pathlength in the context of online convex optimization with switching costs.. In recent work, \cite{baby2019online} and \cite{raj2020non} give dynamic regret bounds for online regression in terms of the total variation of the underlying sequence being estimated. In this paper, we derive the online control policy whose dynamic regret has optimal dependence on the energy of the disturbance $w$, namely $\sum_{t=0}^{T-1} \|w_t\|_2^2$. We call the resulting controller the ``regret-optimal" controller.

\subsection{Contributions of this paper}
Our main result is a derivation of the regret-optimal controller in state-space form using a novel reduction to $H_{\infty}$ control (Theorem \ref{regret-optimal-controller-thm}). Given an $n$-dimensional linear control system and a corresponding LQR cost functional, we show how to construct a $2n$-dimensional linear system and a new cost functional such that the $H_{\infty}$-optimal controller in the new system minimizes dynamic regret in the original system. We show that the regret-optimal controller has a surprising symmetry relative to the offline optimal controller: the regret-optimal control in every round is the sum of the $H_2$-optimal action in the new system and a linear combination of past disturbances, whereas the offline optimal control in every round is the sum of the $H_2$-optimal action in the original system and a linear combination of \textit{future} disturbances.

We emphasize several distinctions between our results and prior work on regret minimization in online control. First, the control policy we derive is the first which is optimal as measured by dynamic regret. In other words, we focus on obtaining an online control policy which competes against the optimal noncausal policy \textit{in the infinite-dimensional set of all possible policies, including time-varying, nonlinear and discontinuous policies}. This stands in stark contrast to all prior work that we are aware of, which focuses on minimizing regret against the optimal fixed policy in some finite-dimensional, parametric class of policies such as state-feedback or disturbance-action policies.
Our results also hold in more general settings than is usually considered in prior work; for example, we allow both the control costs and the dynamics to vary arbitrarily over time. Lastly, the data-dependent regret bound we obtain has a tight dependence on the energy of the disturbance, all the way up to leading constants; while most prior work focuses on finding algorithms with ``order-optimal" regret bounds, we obtain the controller which \textit{exactly} minimizes dynamic regret. The optimality of our controller follows from our reduction to $H_{\infty}$ control; since the variational problem of obtaining a controller with optimal $H_{\infty}$ gain can be solved exactly, the controller we obtain also achieves the exact optimal regret. We also present numerical experiments which show that our regret-optimal controller interpolates between the performance of the $H_2$-optimal and $H_{\infty}$-optimal controllers across stochastic and adversarial environments (Section \ref{numerical-experiments-sec}).  

We extend our analysis to settings where the controller has access to predictions of future disturbances (model-predictive control) and to settings where the controller only affects the system dynamics after a fixed delay. We completely characterize the regret-optimal model predictive controller (Theorem \ref{regret-optimal-mpc-theorem}); our approach is to show that the problem of deriving the regret-optimal controller with predictions can be reduced to the standard regret-optimal control problem without predictions. We show an analogous reduction for control systems with delay and derive the corresponding regret-optimal policy (Theorem \ref{regret-optimal-delay-theorem}). To the best of our knowledge, these policies are the first with optimal dynamic regret in either setting.


\subsection{Related work}
This paper sits at the intersection of two large bodies of work in online learning, the first concerning regret minimization in the control, and the second concerning online learning with dynamic comparators. We briefly summarize some connections to our results.

\vspace{5mm}

\noindent \textbf{Regret minimization in control.} Regret minimization in control has has attracted much recent attention across several distinct settings. A series of papers (e.g. \cite{abbasi2011regret, dean2018regret, cohen2019learning}) consider a model where a fully observed linear system with unknown dynamics (but known costs) is perturbed by stochastic disturbances; the learner picks control actions with the goal of minimizing regret against the class of state-feedback policies. We note that \cite{lale2020regret} considered a similar setting, but with only partial observability. In the ``Non-stochastic control" setting proposed in \cite{agarwal2019online}, the learner knows the dynamics, but the disturbance may be generated adversarially; the controller seeks to minimize regret against the class of disturbance-action policies. An $O(\sqrt{T})$ regret bound was given in \cite{agarwal2019online}; this was recently improved to $O(\log{T})$ in \cite{foster2020logarithmic}.
We emphasize that all of these works focus on minimizing regret against a fixed controller from some parametric class of control policies (policy regret), while we focus on minimizing regret against the optimal dynamic sequence of control actions (dynamic regret).

\vspace{5mm}

\noindent \textbf{Online learning against dynamic comparators.} 
Several papers consider the problem of designing online learning algorithms which compete against dynamic comparators, e.g. \cite{herbster1998tracking, bousquet2002tracking, jadbabaie2015online, goel2019beyond}. The notion of dynamic regret was introduced in \cite{zinkevich2003online};  adaptive regret was introduced in \cite{hazan2009efficient}. A series of papers \cite{besbes2015non, wei2018more} and \cite{bubeck2019improved} give dynamic regret bounds for bandit optimization in terms of the pathlength $\sum_{t=}^{T} \| \ell_t - \ell_{t-1}\|$, which measures the variation in the sequence of losses $\ell_1, \ldots \ell_T$; \cite{goel2019online} also give a dynamic regret bound in terms of pathlength in the context of online convex optimization with switching costs. In recent work, \cite{baby2019online} and \cite{raj2020non} give dynamic regret bounds for online regression in terms of the total variation of the underlying sequence being estimated.


\vspace{5mm}

\noindent \textbf{Control in changing environments.} A key distinction between this paper and most previous work on regret minimization in online control is that we focus on designing an online controller which competes against the optimal dynamic sequence of control actions, instead of the best fixed policy. This problem was first studied in \cite{goel2017thinking} in the context of timescale separation in control (albeit through the lens of competitive ratio rather than regret). In \cite{goel2019online} it was shown that the Online Balanced Descent algorithm introduced in \cite{chen2018smoothed} could be used to give some competitive ratio guarantees in the LQR setting; this result was extended in a series of recent papers, e.g. \cite{goel2019beyond, lin2020online, shi2020online}. A separate line of papers prove dynamic regret bounds for model-predictive control algorithms in settings without dynamics, e.g. \cite{chen2016using, li2018using, li2020leveraging}.
We note that \cite{gradu2020adaptive} give a control algorithm with low adaptive regret against the class of disturbance-action policies. 

\section{Preliminaries}
We consider a linear dynamical system governed by the following evolution equation: 
\begin{equation} \label{evolution-eq}
x_{t+1} = A_tx_t + B_{u, t} u_t + B_{w, t} w_t.
\end{equation}
Here $x_t \in \mathbb{R}^n$ is a state variable we seek to regulate, $u_t \in \mathbb{R}^m$ is a control variable which we can dynamically adjust to influence the evolution of the system, and $w_t \in \mathbb{R}^p$ is an external disturbance. We assume for simplicity that the initial state is $x_0 = 0$. We consider the evolution of this system over a finite time horizon $t = 0, \ldots T-1$ and often use the notation $u = (u_0, \ldots u_{T-1})$, $x = (x_0, \ldots x_{T-1})$, $ w = (w_0, \ldots w_{T-1})$. We formulate the problem of controlling the system as an optimization problem, where the goal is to select the control actions so as to minimize the LQR cost 
\begin{equation} \label{lqr-cost}
\cost(w, u) =  \sum_{t=0}^{T-1} \left( x_t^{\top}Q_t x_t + u_t^{\top}R_t u_t \right), 
\end{equation}
where $Q_t \succeq 0, R_t \succ 0$ for $t = 0, \ldots T - 1$ represent state and control costs, respectively, and $Q_T \succeq 0$ is a terminal state cost.  We assume that the dynamics $\{A_t, B_{u, t}, B_{w, t}\}$ and costs $\{ Q_t, R_t\}_{t=0}^{T-1}$ are known, so the only uncertainty in the evolution of the system comes from the disturbance $w$. We define the \textit{energy} in a signal $w = (w_0, \ldots w_{T-1})$ to be $\|w\|_2^2 = \sum_{t = 0}^{T-1} \|w_t\|^2$.

We say that a control policy is \textit{causal} if the control action it selects at time $t$ depends only on disturbances $w_0, w_1, \ldots, w_t$; otherwise we say the controller is \textit{noncausal}. We often think of controllers as \textit{linear operators} mapping disturbances $w$ to control actions $u$; in this framework, a causal controller is precisely one whose associated operator is causal, i.e. lower-triangular.

For notational convenience, we assume that $R_t = I$ for $t = 0, \ldots T-1$; we emphasize that this imposes no real restriction, since for all $R_t \succ 0$ we can always rescale $u_t$ so that $R_t = I$. More precisely, we can define $B'_{u, t} = B_{u, t}R^{-1/2}_t$ and $u_t' = R^{1/2}u_t$; with this reparameterization, the evolution equation (\ref{evolution-eq}) becomes $$x_{t+1} = A_tx_t + B_{u, t}' u_t' + B_{w, t} w_t,$$ while the state costs $\{Q_t\}_{t=0}^{T-1}$ appearing in (\ref{lqr-cost}) remain unchanged and the control costs are all equal to the identity. We define $s_t = Q^{1/2}_t x_t$ and let $s = (s_0, \ldots s_{T-1})$. Notice that with this notation, the LQR cost (\ref{lqr-cost}) takes a very simple form: 
\begin{equation} \label{lqr-cost2}
\cost(w, u) = \|s\|_2^2 + \|u\|_2^2.
\end{equation}
It is easy to see $s$ can be expressed as a linear combination of $w$ and $u$, i.e. $s = Fu + Gw$, where $F$ and $G$ are strictly causal operators capturing the linear dynamics (\ref{evolution-eq}).

\subsection{State-space models vs input-output models}
A key idea that underpins the results of this paper is that there are two different approaches to describing control policies. The first, and perhaps better-known, approach is to use $\textit{state-space models}$. In this formulation, we describe a control policy explicitly in terms of the quantities appearing in the dynamics $(\ref{evolution-eq})$, such as the state $x_t$, the disturbances $w_0, \ldots w_t$ and the matrices $\{A_t, B_{u, t}, B_{w, t}, Q_t, R_t\}_{t=0}^{T-1}$.  For example, here is a state-space description of the $H_2$-optimal controller obtained by Kalman in his seminal paper \cite{kalman1960contributions}:

\begin{theorem}[Kalman, 1960]  \label{h2-optimal-control-theorem}
The $H_2$-optimal controller has the state-space form $$u_t = -H_t^{-1} B_{u, t}^{\top}P_{t+1}(A_tx_t + B_{w, t}w_t),$$ where $H_t = (R_t + B_{u, t}^{\top}P_{t+1}B_{u, t})$ and $P_t$ is the solution of the backwards Riccati recursion 
\begin{equation} \label{Riccati-recur}
P_{t} =   Q_t + A_t^{\top} P_{t+1}A - A^{\top} P_{t+1}B_t H_t^{-1} P_{t+1} A_t 
\end{equation}
where we initialize $P_T = 0$. 
\end{theorem}

We emphasize that state-space models give a \textit{local} description of the controller; they describe how to compute each control action $u_t$ individually.

The second approach to describe a control policy is to use \textit{input-output models}. In this formulation, we think of a controller $\mathcal{K}$ as an operator mapping a disturbance $w = (w_0, \ldots, w_{T-1})$ to a sequence of control actions $u = (u_0, \ldots, u_{T-1})$, defined in terms of the dynamics encoded by $F$ and $G$. For example, the $H_2$-optimal controller described in Theorem \ref{h2-optimal-control-theorem} has an equivalent input-output description: 

\begin{theorem}[Theorem 11.2.2 in \cite{hassibi1999indefinite}] \label{h2-operator-thm}
The $H_2$-optimal controller $\mathcal{K}$ is given in input-output form as $$\mathcal{K} = -(I + F^{\top}F)^{-1}\{ \Delta^{-1} F^{\top}G\}_{+},$$
where $\{ \mathcal{A} \}_{+}$ denotes the causal part of an operator $\mathcal{A}$ and $\Delta$ is a causal and invertible operator such that $\Delta^{\top}\Delta = I + F^{\top}F$.
\end{theorem}

Input-output models, in contrast to state-space models, give a \textit{global} description of a controller: they describe how to compute the full sequence $u = (u_0, \ldots u_{T-1})$ all at once, without focusing on any particular timestep. We emphasize that input-output models generally do not give a computationally efficient description of a controller. One could, in principle, obtain the $H_{2}$-optimal controller using the formula in Theorem \ref{h2-operator-thm}, but notice that this would take $O(T^3)$ time, since it involves matrix multiplications of $T \times T$ block matrices. By contrast, the state-space controller in Theorem \ref{h2-optimal-control-theorem} can be implemented to run in $O(T)$ time. The main advantage of input-output models is that they are very amenable to algebraic manipulation; in this paper, we first obtain the regret-optimal controller in input-output form and then give more a explicit and computationally efficient description in state-space form.

\subsection{Robust control}
Our result rely heavily on techniques from robust control; in particular, we show that the problem of finding the regret-optimal controller can be encoded as an $H_{\infty}$ control problem:

\begin{problem} [$H_{\infty}$-optimal control problem] \label{hinf-optimal-control-problem}
Find a causal control policy that minimizes $$\sup_{w} \frac{\cost(w, u)}{\|w\|_2^2}. $$
\end{problem}
This problem has the natural interpretation of minimizing the worst-case gain from the energy in the disturbance $w$ to the cost incurred by the controller. In general, it is not known how to derive a closed-form for this $H_{\infty}$ gain or the optimal $H_{\infty}$ controller, so instead is it common to consider a relaxation: 

\begin{problem} [Suboptimal $H_{\infty}$ control problem] \label{hinf-suboptimal-control-problem}
Given a  performance level $\gamma > 0$, find a causal control policy such that  $$\sup_{w} \frac{\cost(w, u)}{\|w\|_2^2} < \gamma^2,  $$ or determine whether no such policy exists.
\end{problem}

This problem has a well-known state-space solution:


\begin{theorem}[Theorem 9.5.1 in \cite{hassibi1999indefinite}] \label{hinf-optimal-thm}
A suboptimal $H_{\infty}$ controller exists at performance level $\gamma$ if and only if
 $$ -\gamma^2I + B_{w, t}^{\top}P_{t+1}B_{w, t} - B_{w, t}^{\top}P_{t+1}B_{u, t} H_t^{-1}B_{u, t}^{\top}P_{t+1}B_{w, t} \prec 0 $$
for all $t = 0, \ldots T-1$, where we define
$$H_t = R_t + B_{u, t}^{\top}P_{t+1}B_{u, t},$$
$P_t$ is the solution of the backwards-time Riccati equation $$P_t = Q_t + A_t^{\top}P_{t+1}A_t -  A_t^{\top}P_{t+1}\hat{B}_t \hat{H}_t^{-1} \hat{B}_t^{\top} P_{t+1}A_t $$ with initialization $P_T = Q_T$, and we define $$ \hat{B}_t = \begin{bmatrix} B_{u, t} & B_{w, t} \end{bmatrix}, $$ $$\hat{H}_t = \begin{bmatrix} R_t & 0 \\ 0 & -\gamma^2 I \end{bmatrix}  + \hat{B}_t^{\top} P_{t+1} \hat{B}_t.$$
In this case, the suboptimal $H_{\infty}$ controller has the form $$ u_t =  -H_t^{-1} B_{u, t}^{\top}P_{t+1}(A_tx_t +  B_{w, t}  w_t).$$
\end{theorem}

Note that the $H_{\infty}$-optimal controller described in Problem (\ref{hinf-optimal-control-problem}) is easily obtained from the solution of the suboptimal $H_{\infty}$ problem by bisection on $\gamma$. We iteratively reduce the value of $\gamma$ until we find the smallest value of $\gamma$ such that the constraints described in Theorem \ref{hinf-optimal-thm} are satisfied; the corresponding controller is the optimal $H_{\infty}$ controller (we refer the reader to  \cite{doyle1988state, uchida1990finite, nagpal1991filtering, limebeer1989discrete} for details).

\subsection{Regret-optimal control}
In this paper, instead of minimizing the worst-case \textit{cost} as in $H_{\infty}$ control, our goal is to minimize the worst-case \textit{regret}.  This problem has a natural analog of the $H_{\infty}$ problem:

\begin{problem} [Regret-optimal control problem] \label{regret-optimal-control-problem}
Find a causal control policy that minimizes $$\sup_{w} \frac{\cost(w, u) - \min_{u} \cost(w, u)}{\|w\|_2^2}.  $$
\end{problem}

\noindent This problem has the natural interpretation of minimizing the worst-case gain from the energy in the disturbance $w$ to the regret incurred by the controller against the optimal dynamic sequence of control actions. As in the $H_{\infty}$ setting, we consider the relaxation:

\begin{problem} [Regret-suboptimal control problem] \label{regret-suboptimal-control-problem}
Given a performance level $\gamma > 0$, find a causal control policy such that $$ \frac{\cost(w, u) - \min_{u} \cost(w, u) }{ \|w\|_2^2} < \gamma^2  $$ for all disturbances $w$, or determine whether no such policy exists.
\end{problem}

\noindent Note that if we can find a regret-suboptimal controller at some performance level $\gamma$, we immediately obtain a data-dependent bound on its dynamic regret: $$\cost(w, u) - \min_{u} \cost(w, u) < \gamma \|w\|_2^2.$$

\noindent We emphasize that, as in the $H_{\infty}$ setting, if we can solve the regret-suboptimal  problem, we can easily recover the solution to the regret-optimal problem by bisection on $\gamma$.

\subsection{The optimal noncausal controller}
We briefly state a few facts about the optimal noncausal controller (sometimes called the offline optimal controller), i.e. the controller which selects the optimal dynamic sequence of control actions given full knowledge of the disturbance $w$: $$u^* = \argmin \cost(w, u).$$ Recall from equation (\ref{lqr-cost2}) that $$\cost(w, u) = \|Fu + Gw\|_2^2 + \|w\|^2.$$ Solving for the cost-minimizing $u$ via completion-of-squares, we see that
\begin{equation}  \label{offline-operator}
u^* = - (I + F^{\top}F)^{-1}F^{\top}Gw.
\end{equation}
and
\begin{equation} \label{offline-operator-cost}
\cost(w; u^*) = w^{\top}G^{\top}(I + FF^{\top})^{-1}Gw.
\end{equation}
While equation (\ref{offline-operator-cost}) shows that the optimal noncausal control action is a linear function of the disturbance $w$, we emphasize that we did not impose any structure on the optimal noncausal controller; $u^*$ is the optimal choice out of all possible functions of $w$, including nonlinear and even discontinuous functions of $w$. Equation (\ref{offline-operator}) gives the optimal noncausal controller in input-output form; recently \cite{goel2020power} and \cite{foster2020logarithmic} obtained a state-space description using dynamic programming:

\begin{theorem}[\cite{goel2020power}] \label{offline-structure-thm}
The optimal noncausal controller has the form $$u_t^* = -H_t^{-1}B_{u, t}^{\top} \left(P_{t+1}A_t x_t + P_{t+1}B_{w, t} w_t + \frac{1}{2}v_{t+1} \right),$$ where we define $$H_t = R_t + B_{u, t}^{\top}P_{t+1}B_{u, t}, $$ $P_t$ is the solution of the Riccati recurrence (\ref{Riccati-recur}), and $v_t$ satisfies the recurrence 
$$ v_t = 2A_t^{\top}S_t B_{w, t} w_t + A_t^{\top}S_tP_{t+1}^{-1}v_{t+1}, $$
and we define 
$$S_t = P_{t+1} - P_{t+1}B_{u, t}H_t^{-1}B_{u, t}^{\top}P_{t+1}.$$
\end{theorem}
In light of Theorem \ref{h2-optimal-control-theorem}, this shows that \textit{the regret-optimal control action at time $t$ is the sum of the $H_2$-optimal control action and a linear combination of the future disturbances $w_{t+1}, \ldots w_{T-1}$.}

\section{The regret-optimal controller} \label{regret-optimal-sec}
We now turn to the problem of deriving the regret-optimal controller. Our approach is to reduce the  regret-suboptimal control problem (Problem \ref{regret-suboptimal-control-problem})  to the suboptimal $H_{\infty}$ problem (Problem \ref{hinf-suboptimal-control-problem}). As in the $H_{\infty}$ setting, once the regret-suboptimal controller is found, the regret-optimal controller is easily obtained by bisection.

Recall that the regret-suboptimal problem (Problem \ref{regret-suboptimal-control-problem}) with performance level $\gamma$ is to find, if possible, a causal control policy such that for all disturbances $w$, 
$$\cost(w, u) - \min_{u} \cost(w, u)  < \gamma^2 \|w\|_2^2. $$
In light of equations (\ref{lqr-cost2}) and (\ref{offline-operator-cost}), this condition can be rewritten as 
\begin{equation} \label{regret-suboptimal-problem-2}
\|u\|_2^2 + \|s\|_2^2 < w^{\top} (\gamma^2 I +  G^{\top}(I + FF^{\top})^{-1}G)w,
\end{equation}
where $s = Fu + Gw$.
Our approach shall be to find a change of variables such that this problem takes the form of the suboptimal $H_{\infty}$ control problem (Problem \ref{hinf-suboptimal-control-problem}). Since $\gamma^2 I +  G^{\top}(I + FF^{\top})^{-1}G$ is strictly positive definite, there exists a unique causal, invertible matrix $\Delta$ such that $$\gamma^2 I +  G^{\top}(I + FF^{\top})^{-1}G = \Delta^{\top}\Delta.$$ Notice that $$w^{\top} (\gamma^2 I +  G^{\top}(I + FF^{\top})^{-1}G)w = \|\Delta w\|_2^2.$$ Letting $z = \Delta w$ and $G' = G\Delta^{-1}$, we have $s = Fu + G' z$. With this change of variables, the regret-suboptimal problem (\ref{regret-suboptimal-problem-2}) takes the form of finding, if possible, a causal control policy such that for all $w$, $$\|u\|_2^2 + \|s\|_2^2 < \|z\|_2^2.$$
Notice that this is a suboptimal $H_{\infty}$ problem. We have shown:

\begin{theorem} \label{reduction-theorem}
Consider a linear dynamical system with dynamics given by $F$ and $G$. A causal controller $\mathcal{K}$ is a solution to the regret-suboptimal problem (Problem \ref{regret-suboptimal-control-problem}) in this system at performance level $\gamma$ if and only if $\mathcal{K}$ is a solution to the suboptimal $H_{\infty}$  problem  (Problem \ref{hinf-suboptimal-control-problem}) with performance level 1 in a linear dynamical system with dynamics given by $F$ and $G'$, where $G' = G\Delta^{-1}$ and $\Delta$ is the unique causal operator such that $$\gamma^2 I +  G^{\top}(I + FF^{\top})^{-1}G = \Delta^{\top}\Delta.$$
\end{theorem}

Using Theorem \ref{reduction-theorem}, we can easily obtain the regret-optimal controller in input-output form; we refer the reader to Section \ref{regret-suboptimal-inputoutput-sec} for details and concentrate on obtaining the regret-optimal controller in state-space form.

The main technical challenge is to obtain an explicit factorization of $\gamma^2 + G^{\top}(I + F F^{\top})^{-1}G$ as $\Delta^{\top}\Delta$; once we have obtained $\Delta$ it is straightforward to recover a state-space description of the regret-optimal controller using the state-space description of the $H_{\infty}$-optimal controller (Theorem \ref{hinf-optimal-thm}). We briefly give a high-level overview of our proof technique. Our approach is centered around repeated use of the celebrated Kalman filter, which, given a state-space model for a random variable whose covariance is a positive-definite operator $\Sigma$, produces a causal state-space model for an operator $M$ such that $\Sigma = M^{\top}M$. Alternatively, we can use the backwards Kalman filter (i.e. the standard Kalman filter given observations in reverse order) to obtain a factorization $MM^{\top}$, where $M$ is causal. We first construct a state-space model representing a random variable with covariance $I + FF^{\top}$; we then use the Kalman filter to factor this operator as $LL^{\top}$ where $L$ is causal. Using the state-space model for $L$, we construct a new random variable with covariance $\gamma^2I + G^{\top}(I + F F^{\top})G$ and use the backwards Kalman filter to factor this operator as $\Delta^{\top}\Delta$ where $\Delta$ is causal.

We state our main result, and defer the technical details to Section \ref{regret-optimal-appendix-sec}.

\begin{theorem} \label{regret-optimal-controller-thm}
The regret-suboptimal controller at performance level $\gamma$ has the form
$$ u_t = - \hat{H}^{-1}_t \hat{B}_{u, t}^{\top} \hat{P}_{t+1} \left(\hat{A}_t \begin{bmatrix} \zeta_t \\ \hat{\nu}_t \end{bmatrix} + \hat{B}_{w, t} z_t \right), $$
 where we define 
$$\hat{A}_t = \begin{bmatrix} A_t & - B_{w, t}(K^b_{l, t})^{\top} \\ 0 &  \tilde{A}_t - B_{w, t}(K^b_{l, t})^{\top} \end{bmatrix}, \hspace{5mm} \hat{B}_{u, t} =  \begin{bmatrix} B_{u, t} \\ 0 \end{bmatrix}, \hspace{5mm} \hat{B}_{w, t} =  \begin{bmatrix} B_{w, t} (R^b_{e, t})^{-\frac{1}{2}} \\ B_{w, t} (R^b_{e, t})^{-\frac{1}{2}} \end{bmatrix}, $$
$$\hat{Q}_t = \begin{bmatrix} Q_t & 0 \\ 0 & 0 \end{bmatrix}, \hspace{5mm} \hat{H}_t = I + \hat{B}_{u, t}^{\top} \hat{P}_{t+1} \hat{B}_{u, t}, \hspace{5mm} \tilde{A}_t = A_t - K_{p, t} Q_t^{\frac{1}{2}}, $$
$$K_{p, t} = A_t P_t Q_t^{\frac{1}{2}} R_{e, t}^{-1}, \hspace{5mm} R_{e, t} = I + Q_t^{\frac{1}{2}}P_t Q_t^{\frac{1}{2}}, \hspace{5mm} K_{l, t}^b = \tilde{A}_t^{\top}P_t^b B_{w, t} (R_{e, t}^b)^{-1}, $$ 
$$R_{e, t}^b = \gamma^2 I + B^{\top}_{w, t} P_t^b B_{w, t}, \hspace{5mm} \delta_{t+1} =  \tilde{A}_t\delta_t + B_{w, t} w_t, \hspace{5mm}  z_t = (R_{e, t}^b)^{\frac{1}{2}} (K_{l, t}^b)^{\top}\delta_t  +   (R_{e,t}^b)^{\frac{1}{2}}w_t, $$
and we initialize $\delta_0 = 0$. The state variables $\zeta_t$ and $\hat{\nu}_t$ evolve according to to the dynamics
$$ \begin{bmatrix} \zeta_{t+1} \\ \hat{\nu}_{t+1} \end{bmatrix} = \hat{A}_t \begin{bmatrix} \zeta_t \\ \hat{\nu}_t \end{bmatrix} + \hat{B}_{u, t} u_t + \hat{B}_{w, t} z_t, $$
where we initialize $\zeta_0, \hat{\nu}_0 = 0$. We define 
 $P_t$ to be the solution of the forwards Riccati recursion
 $$P_{t+1} = A_tP_tA_t^{\top} + B_{u, t}B_{u, t}^{\top} - K_{p, t} R_{e, t}K_{p, t}^{\top}, $$ where we initialize $P_0 = 0$, and $\hat{P}_t, P_t^b$ to be the solutions of the backwards Riccati recursions 
$$ P_{t-1}^b =  \tilde{A}_t^{\top} P_{t}^b \tilde{A}_t + Q_t^{\frac{1}{2}}(R_{e, t})^{-1}Q_t^{\frac{1}{2}} - K_{l, t}^b R_{e, t}^b (K_{l, t}^b)^{\top}, $$
$$\hat{P}_{t} = \hat{Q}_t + \hat{A}_t^{\top}\hat{P}_{t+1}A_t - \hat{A}_t^{\top} \hat{P}_{t+1} \hat{B}_{u, t} \hat{H}_t^{-1} \hat{B}_{u, t}^{\top} \hat{P}_{t+1} \hat{A}_t, $$ where we initialize $P^b_T = 0, \hat{P}_T = 0$. The regret-optimal controller is the regret-suboptimal controller at performance level $\gamma_{opt}$, where $\gamma_{opt}$ is the smallest value of $\gamma$ such that  $$ -\gamma^2 I +  \hat{B}_{w, t}^{\top} \hat{P}_{t+1} \hat{B}_{w, t} - \hat{B}_{w, t}^{\top} \hat{P}_{t+1}\hat{B}_{u, t} \hat{H}_t^{-1}\hat{B}_{u, t}^{\top}\hat{P}_{t+1} \hat{B}_{w, t} \prec 0$$ for all $t = 0, \ldots T-1$. Furthermore, the regret incurred by the regret-optimal controller on any specific disturbance $w = (w_0,  \ldots w_{T-1})$ is at most $\gamma_{opt}^2 \|w\|_2^2.$
 \end{theorem}

\noindent The regret-optimal controller described in Theorem \ref{regret-optimal-controller-thm} may appear mysterious, so we take a brief detour to better understand its structure. We introduce the block decomposition $$\hat{P}_t = \begin{bmatrix} P_{11, t} & P_{12, t} \\ P_{21, t} & P_{22, t} \end{bmatrix}, $$ where each of the  submatrices has size $n \times n$. With this notation, we see that $$\hat{H}_t = I + B_{u, t}^{\top}P_{11, t} B_{u, t}$$ and
\begin{align*}
 u_t &= -\hat{H}^{-1}_t B_{u, t}^{\top} \begin{bmatrix} P_{11, t} &
 P_{12, t}  \end{bmatrix} \left(\hat{A}_t
 \begin{bmatrix} \zeta_t \\ \hat{\nu}_t \end{bmatrix} + \hat{B}_{w, t} z_t \right) \\
 &= -\hat{H}_t^{-1} B_{u, t}^{\top} P_{11, t}(A_t \zeta_t + B_{w, t} (R_{l, t}^b)^{-\frac{1}{2}} z_t)  
  + \hat{H}_t^{-1} B_{u, t}^{\top} P_{11, t} B_{w, t}(K^b_{l, t})^{\top} \hat{\nu}_t \\
  & - \hat{H}_t^{-1} B_{u, t}^{\top}  P_{12, t} (\tilde{A}_t - B_{w, t}(K^b_{l, t})^{\top}) \hat{\nu}_t,
\end{align*}
where we repeatedly used the fact that the (2, 1) block submatrix of $\hat{B}_{u, t}$ is zero.
After simplifying the backwards Riccati recursion for $\hat{P}_t$, we see that $P_{11, t}$ satisfies the recursion
$$P_{11, t} =   Q_t + A_t^{\top} P_{11, t+1}A - A^{\top} P_{11, t+1}B_t \hat{H}_t^{-1} P_{11, t+1} A_t, $$ where $P_{11, T} = 0$. Notice that this is precisely the recursion (\ref{Riccati-recur}) that appears in the $H_2$-optimal controller. We can hence recognize the term $-\hat{H}_t^{-1} B_{u, t}^{\top} P_{11, t}A_t \zeta_t$ as the optimal $H_2$ control action described in Theorem \ref{h2-optimal-control-theorem}. It is worth emphasizing this result: 

\vspace{2mm}
\textit{The regret-optimal control action at time $t$ is the sum of the $H_2$-optimal control action and a linear combination of past disturbances $w_0, \ldots w_t$.}
\vspace{2mm}

\noindent This result is especially interesting in light of Theorem \ref{offline-structure-thm}, which shows that the optimal noncausal control action is the sum of $H_2$-optimal control action and a linear combination of \textit{future} disturbances.

\subsection{Regret-optimal control with predictions and delay}
Our techniques easily extend to settings where the controller can predict the next $h$ disturbances (model-predictive control) and to settings where the controller is only able to affect the system dynamics after a delay of $d$ timesteps; we refer the reader to Section \ref{predictions-delay-sec} for details.

\section{Numerical experiments}
We benchmark the $H_2$-optimal, $H_{\infty}$-optimal, optimal noncausal, and regret-optimal controllers in the context of the inverted pendulum, a classic control system which has been widely studied in works at the intersection of machine learning and control. Our experiments show that our regret-optimal controller interpolates between the performance of the $H_2$-optimal and $H_{\infty}$-optimal controllers across stochastic and adversarial environments; we refer the reader to Section \ref{numerical-experiments-sec} for details and plots. 

\section{Conclusion}
In this paper, we consider regret minimization in online control through the lens of dynamic regret, a more challenging benchmark than policy regret which is studied in much prior work. We obtain a computationally efficient state-space description of the regret-optimal controller via a novel reduction to $H_{\infty}$ control, and present a simple bound on its dynamic regret in terms of the energy of the disturbance. We show that our regret-optimal controller has a surprising symmetry relative to the offline optimal controller: the regret-optimal control action in each round is the sum of the action selected by an $H_2$-optimal controller and a linear combination of past disturbances, whereas the offline optimal control in each round is the sum of the action selected by an $H_2$-optimal controller and a linear combination of \textit{future} disturbances. 

We hope that this work will help bring operator-theoretic tools from robust control and signal processing into the machine learning community. In future work, we plan to extend the results of this paper to the more challenging \textit{partial observability} setting, where the online policy only has access to noisy linear estimates of the state. It would also be interesting to consider systems with nonlinear dynamics and measure the performance of a receding-horizon control policy which iteratively linearizes the system dynamics and selects the corresponding regret-optimal controller.

\newpage
\bibliography{main}

\begin{thebibliography}{37}
\providecommand{\natexlab}[1]{#1}
\providecommand{\url}[1]{\texttt{#1}}
\expandafter\ifx\csname urlstyle\endcsname\relax
  \providecommand{\doi}[1]{doi: #1}\else
  \providecommand{\doi}{doi: \begingroup \urlstyle{rm}\Url}\fi

\bibitem[Abbasi-Yadkori and Szepesvari(2011)]{abbasi2011regret}
Yasin Abbasi-Yadkori and Csaba Szepesvari.
\newblock Regret bounds for the adaptive control of linear quadratic systems.
\newblock In \emph{Proceedings of the 24th Annual Conference on Learning
  Theory}, pages 1--26, 2011.

\bibitem[Agarwal et~al.(2019)Agarwal, Bullins, Hazan, Kakade, and
  Singh]{agarwal2019online}
Naman Agarwal, Brian Bullins, Elad Hazan, Sham~M Kakade, and Karan Singh.
\newblock Online control with adversarial disturbances.
\newblock \emph{arXiv preprint arXiv:1902.08721}, 2019.

\bibitem[{\AA}str{\"o}m and Murray(2021)]{aastrom2021feedback}
Karl~Johan {\AA}str{\"o}m and Richard~M Murray.
\newblock \emph{Feedback systems: an introduction for scientists and
  engineers}.
\newblock Princeton university press, 2021.

\bibitem[Baby and Wang(2019)]{baby2019online}
Dheeraj Baby and Yu-Xiang Wang.
\newblock Online forecasting of total-variation-bounded sequences.
\newblock \emph{arXiv preprint arXiv:1906.03364}, 2019.

\bibitem[Besbes et~al.(2015)Besbes, Gur, and Zeevi]{besbes2015non}
Omar Besbes, Yonatan Gur, and Assaf Zeevi.
\newblock Non-stationary stochastic optimization.
\newblock \emph{Operations research}, 63\penalty0 (5):\penalty0 1227--1244,
  2015.

\bibitem[Bousquet and Warmuth(2002)]{bousquet2002tracking}
Olivier Bousquet and Manfred~K Warmuth.
\newblock Tracking a small set of experts by mixing past posteriors.
\newblock \emph{Journal of Machine Learning Research}, 3\penalty0
  (Nov):\penalty0 363--396, 2002.

\bibitem[Bubeck et~al.(2019)Bubeck, Li, Luo, and Wei]{bubeck2019improved}
S{\'e}bastien Bubeck, Yuanzhi Li, Haipeng Luo, and Chen-Yu Wei.
\newblock Improved path-length regret bounds for bandits.
\newblock In \emph{Conference On Learning Theory}, pages 508--528. PMLR, 2019.

\bibitem[Chen et~al.(2016)Chen, Comden, Liu, Gandhi, and
  Wierman]{chen2016using}
Niangjun Chen, Joshua Comden, Zhenhua Liu, Anshul Gandhi, and Adam Wierman.
\newblock Using predictions in online optimization: Looking forward with an eye
  on the past.
\newblock \emph{ACM SIGMETRICS Performance Evaluation Review}, 44\penalty0
  (1):\penalty0 193--206, 2016.

\bibitem[Chen et~al.(2018)Chen, Goel, and Wierman]{chen2018smoothed}
Niangjun Chen, Gautam Goel, and Adam Wierman.
\newblock Smoothed online convex optimization in high dimensions via online
  balanced descent.
\newblock \emph{arXiv preprint arXiv:1803.10366}, 2018.

\bibitem[Cohen et~al.(2019)Cohen, Koren, and Mansour]{cohen2019learning}
Alon Cohen, Tomer Koren, and Yishay Mansour.
\newblock Learning linear-quadratic regulators efficiently with only $\sqrt{T}$
  regret.
\newblock \emph{arXiv preprint arXiv:1902.06223}, 2019.

\bibitem[Dean et~al.(2018)Dean, Mania, Matni, Recht, and Tu]{dean2018regret}
Sarah Dean, Horia Mania, Nikolai Matni, Benjamin Recht, and Stephen Tu.
\newblock Regret bounds for robust adaptive control of the linear quadratic
  regulator.
\newblock In \emph{Advances in Neural Information Processing Systems}, pages
  4188--4197, 2018.

\bibitem[Doyle et~al.(1988)Doyle, Glover, Khargonekar, and
  Francis]{doyle1988state}
John Doyle, Keith Glover, Pramod Khargonekar, and Bruce Francis.
\newblock State-space solutions to standard h 2 and h-infinity control
  problems.
\newblock In \emph{1988 American Control Conference}, pages 1691--1696. IEEE,
  1988.

\bibitem[Doyle(1978)]{doyle1978guaranteed}
John~C Doyle.
\newblock Guaranteed margins for lqg regulators.
\newblock \emph{IEEE Transactions on automatic Control}, 23\penalty0
  (4):\penalty0 756--757, 1978.

\bibitem[Foster and Simchowitz(2020)]{foster2020logarithmic}
Dylan~J Foster and Max Simchowitz.
\newblock Logarithmic regret for adversarial online control.
\newblock \emph{arXiv preprint arXiv:2003.00189}, 2020.

\bibitem[Goel and Hassibi(2020)]{goel2020power}
Gautam Goel and Babak Hassibi.
\newblock The power of linear controllers in lqr control.
\newblock \emph{arXiv preprint arXiv:2002.02574}, 2020.

\bibitem[Goel and Wierman(2019)]{goel2019online}
Gautam Goel and Adam Wierman.
\newblock An online algorithm for smoothed regression and lqr control.
\newblock \emph{Proceedings of Machine Learning Research}, 89:\penalty0
  2504--2513, 2019.

\bibitem[Goel et~al.(2017)Goel, Chen, and Wierman]{goel2017thinking}
Gautam Goel, Niangjun Chen, and Adam Wierman.
\newblock Thinking fast and slow: Optimization decomposition across timescales.
\newblock In \emph{2017 IEEE 56th Annual Conference on Decision and Control
  (CDC)}, pages 1291--1298. IEEE, 2017.

\bibitem[Goel et~al.(2019)Goel, Lin, Sun, and Wierman]{goel2019beyond}
Gautam Goel, Yiheng Lin, Haoyuan Sun, and Adam Wierman.
\newblock Beyond online balanced descent: An optimal algorithm for smoothed
  online optimization.
\newblock In \emph{Advances in Neural Information Processing Systems}, pages
  1875--1885, 2019.

\bibitem[Gradu et~al.(2020)Gradu, Hazan, and Minasyan]{gradu2020adaptive}
Paula Gradu, Elad Hazan, and Edgar Minasyan.
\newblock Adaptive regret for control of time-varying dynamics.
\newblock \emph{arXiv preprint arXiv:2007.04393}, 2020.

\bibitem[Hassibi et~al.(1999)Hassibi, Sayed, and
  Kailath]{hassibi1999indefinite}
Babak Hassibi, Ali~H Sayed, and Thomas Kailath.
\newblock \emph{Indefinite-quadratic estimation and control: a unified approach
  to H 2 and H-infinity theories}.
\newblock SIAM, 1999.

\bibitem[Hazan and Seshadhri(2009)]{hazan2009efficient}
Elad Hazan and Comandur Seshadhri.
\newblock Efficient learning algorithms for changing environments.
\newblock In \emph{Proceedings of the 26th annual international conference on
  machine learning}, pages 393--400, 2009.

\bibitem[Hazan et~al.(2020)Hazan, Kakade, and Singh]{hazan2020nonstochastic}
Elad Hazan, Sham Kakade, and Karan Singh.
\newblock The nonstochastic control problem.
\newblock In \emph{Algorithmic Learning Theory}, pages 408--421. PMLR, 2020.

\bibitem[Herbster and Warmuth(1998)]{herbster1998tracking}
Mark Herbster and Manfred~K Warmuth.
\newblock Tracking the best expert.
\newblock \emph{Machine learning}, 32\penalty0 (2):\penalty0 151--178, 1998.

\bibitem[Jadbabaie et~al.(2015)Jadbabaie, Rakhlin, Shahrampour, and
  Sridharan]{jadbabaie2015online}
Ali Jadbabaie, Alexander Rakhlin, Shahin Shahrampour, and Karthik Sridharan.
\newblock Online optimization: Competing with dynamic comparators.
\newblock In \emph{Artificial Intelligence and Statistics}, pages 398--406.
  PMLR, 2015.

\bibitem[Kailath et~al.(2000)Kailath, Sayed, and Hassibi]{kailath2000linear}
Thomas Kailath, Ali~H Sayed, and Babak Hassibi.
\newblock \emph{Linear estimation}.
\newblock Prentice Hall, 2000.

\bibitem[Kalman et~al.(1960)]{kalman1960contributions}
Rudolf~Emil Kalman et~al.
\newblock Contributions to the theory of optimal control.
\newblock \emph{Bol. soc. mat. mexicana}, 5\penalty0 (2):\penalty0 102--119,
  1960.

\bibitem[Lale et~al.(2020)Lale, Azizzadenesheli, Hassibi, and
  Anandkumar]{lale2020regret}
Sahin Lale, Kamyar Azizzadenesheli, Babak Hassibi, and Anima Anandkumar.
\newblock Regret bound of adaptive control in linear quadratic gaussian (lqg)
  systems.
\newblock \emph{arXiv preprint arXiv:2003.05999}, 2020.

\bibitem[Li and Li(2020)]{li2020leveraging}
Yingying Li and Na~Li.
\newblock Leveraging predictions in smoothed online convex optimization via
  gradient-based algorithms.
\newblock \emph{arXiv preprint arXiv:2011.12539}, 2020.

\bibitem[Li et~al.(2018)Li, Qu, and Li]{li2018using}
Yingying Li, Guannan Qu, and Na~Li.
\newblock Using predictions in online optimization with switching costs: A fast
  algorithm and a fundamental limit.
\newblock In \emph{2018 Annual American Control Conference (ACC)}, pages
  3008--3013. IEEE, 2018.

\bibitem[Limebeer et~al.(1989)Limebeer, Green, and
  Walker]{limebeer1989discrete}
DJN Limebeer, Michael Green, and D~Walker.
\newblock Discrete-time h/sup infinity/control.
\newblock In \emph{Proceedings of the 28th IEEE Conference on Decision and
  Control,}, pages 392--396. IEEE, 1989.

\bibitem[Lin et~al.(2020)Lin, Goel, and Wierman]{lin2020online}
Yiheng Lin, Gautam Goel, and Adam Wierman.
\newblock Online optimization with predictions and non-convex losses.
\newblock \emph{Proceedings of the ACM on Measurement and Analysis of Computing
  Systems}, 4\penalty0 (1):\penalty0 1--32, 2020.

\bibitem[Nagpal and Khargonekar(1991)]{nagpal1991filtering}
Krishan~M Nagpal and Pramod~P Khargonekar.
\newblock Filtering and smoothing in an h/sup infinity/setting.
\newblock \emph{IEEE Transactions on Automatic Control}, 36\penalty0
  (2):\penalty0 152--166, 1991.

\bibitem[Raj et~al.(2020)Raj, Gaillard, and Saad]{raj2020non}
Anant Raj, Pierre Gaillard, and Christophe Saad.
\newblock Non-stationary online regression.
\newblock \emph{arXiv preprint arXiv:2011.06957}, 2020.

\bibitem[Shi et~al.(2020)Shi, Lin, Chung, Yue, and Wierman]{shi2020online}
Guanya Shi, Yiheng Lin, Soon-Jo Chung, Yisong Yue, and Adam Wierman.
\newblock Online optimization with memory and competitive control.
\newblock \emph{arXiv e-prints}, pages arXiv--2002, 2020.

\bibitem[Uchida and Fujita(1990)]{uchida1990finite}
Kenko Uchida and Mayasuki Fujita.
\newblock Finite horizon h/sup infinity/control problems with terminal
  penalties.
\newblock In \emph{29th IEEE Conference on Decision and Control}, pages
  1808--1813. IEEE, 1990.

\bibitem[Wei and Luo(2018)]{wei2018more}
Chen-Yu Wei and Haipeng Luo.
\newblock More adaptive algorithms for adversarial bandits.
\newblock In \emph{Conference On Learning Theory}, pages 1263--1291. PMLR,
  2018.

\bibitem[Zinkevich(2003)]{zinkevich2003online}
Martin Zinkevich.
\newblock Online convex programming and generalized infinitesimal gradient
  ascent.
\newblock In \emph{Proceedings of the 20th international conference on machine
  learning (icml-03)}, pages 928--936, 2003.

\end{thebibliography}

\newpage 
\appendix

\section{The regret-optimal controller in input-output form} \label{regret-suboptimal-inputoutput-sec}

We first describe the suboptimal $H_{\infty}$ controller in input-output form:

\begin{theorem}[Theorem 11.3.3 in \cite{hassibi1999indefinite}]
A suboptimal $H_{\infty}$ controller $\mathcal{K}$ at performance level $\gamma$ exists if and only if the operator 

$$\begin{bmatrix} I + F^{\top}F & F^{\top}G \\ G^{\top}F & -\gamma^2 I + G^{\top}G \end{bmatrix} $$
can be factored as $$ \begin{bmatrix} L_{11}^{\top} & L_{21}^{\top} \\ L_{12}^{\top} & L_{22}^{\top} \end{bmatrix} \begin{bmatrix} I & 0 \\ 0 & -I \end{bmatrix}  \begin{bmatrix} L_{11} & L_{12} \\ L_{21} & L_{22} \end{bmatrix},$$ where $\begin{bmatrix} L_{11} & L_{12} \\ L_{21} & L_{22} \end{bmatrix}$ is causal and invertible, $L_{11}$ is causal and invertible, and $L_{22}$ is strictly causal. In this case, the suboptimal $H_{\infty}$ controller at performance level $\gamma$ is given by $\mathcal{K} = -L_{11}^{-1}L_{12}$.
\end{theorem}

\noindent In light of Theorem \ref{reduction-theorem}, we immediately obtain an input-output description of the regret-suboptimal controller.

\begin{theorem} \label{regret-suboptimal-inputoutput-theorem}
A regret-suboptimal controller $\mathcal{K}$ at performance level $\gamma$ exists if and only if the operator 

$$\begin{bmatrix} I + F^{\top}F & F^{\top}G\Delta^{-1} \\ \Delta^{-\top}G^{\top}F & -I + \Delta^{-\top}G^{\top}G\Delta^{-1} \end{bmatrix} $$
can be factored as $$ \begin{bmatrix} L_{11}^{\top} & L_{21}^{\top} \\ L_{12}^{\top} & L_{22}^{\top} \end{bmatrix} \begin{bmatrix} I & 0 \\ 0 & -I \end{bmatrix}  \begin{bmatrix} L_{11} & L_{12} \\ L_{21} & L_{22} \end{bmatrix},$$ where $\begin{bmatrix} L_{11} & L_{12} \\ L_{21} & L_{22} \end{bmatrix}$ is causal and invertible, $L_{11}$ is causal and invertible, $L_{22}$ is strictly causal, and we define $\Delta$ to be the unique causal operator such that $$\gamma^2 I +  G^{\top}(I + FF^{\top})^{-1}G = \Delta^{\top}\Delta.$$ In this case, the regret-suboptimal controller at performance level $\gamma$ is given by $\mathcal{K} = -L_{11}^{-1}L_{12}$.
\end{theorem}

\noindent The regret-optimal controller is the regret-suboptimal controller at performance level $\gamma_{opt}$; where $\gamma_{opt}$ is the smallest value of $\gamma$ such that the factorization described in Theorem \ref{regret-suboptimal-inputoutput-theorem} exists; an equivalent description of the regret-optimal controller and $\gamma_{opt}$ is described in Theorem \ref{regret-optimal-controller-thm}. 

\section{Proof of Theorem \ref{regret-optimal-controller-thm}} \label{regret-optimal-appendix-sec}

\begin{proof}
Consider the state-space model $$\xi_{t+1} = A_t \xi_t + B_{u, t} u_t, \hspace{5mm} y_t = Q_t^{\frac{1}{2}} \xi_t + v_t,$$ where $u_t$, $v_t$ are zero mean noise variables such that $\expect[u_t u_t^{\top}] = \expect[v_t v_t^{\top}] = I$ and $\expect[u_t v_t^{\top}] = 0$. Let $y = (y_0, \ldots y_{T-1})$, $u = (u_0, \ldots, u_{T-1})$, and $v = (v_0, \ldots, v_{T-1})$. Notice that $y = Fu + v$ and $\expect[y y^{\top}] = I + F F^{\top}$. Suppose we can find a causal matrix $L$ such that $y = L e$ where $e$ is a zero-mean random variable such that $\expect[e e^{\top}] = I$. Then $\expect[y y^{\top}] = LL^{\top}$, so $I + FF^{\top} = L L^{\top}$. 

Using the Kalman filter (as described in Theorem 9.2.1 in \cite{kailath2000linear}), we obtain a state-space model for $L$ given by 
\begin{equation} \label{first-factorization}
\hat{\xi}_{t+1} = A_t \hat{\xi}_t + K_{p, t} R^{\frac{1}{2}}_{e, t} e_t, \hspace{5mm} y_t = Q_t^{\frac{1}{2}} \hat{\xi}_t +  R^{\frac{1}{2}}_{e, t}e_t,
\end{equation}
where we define $K_{p, t} = A_t P_t Q_t^{\frac{1}{2}} R_{e, t}^{-1}$ and $R_{e, t} = I + Q_t^{\frac{1}{2}}P_t Q_t^{\frac{1}{2}}$ and $P_t$ is defined recursively as $$P_{t+1} = A_tP_tA_t^{\top} + B_{u, t}B_{u, t}^{\top} - K_{p, t} R_{e, t}K_{p, t}^{\top}$$ and $P_0 = 0$. Exchanging inputs and outputs, we see that a state-space model for $L^{-1}$ is $$\hat{\xi}_{t+1} = A_t\hat{\xi}_t + K_{p, t}(y_t - Q_t^{\frac{1}{2}}\hat{\xi}_t), \hspace{5mm} e_t = R^{-\frac{1}{2}}_{e, t}(y_t - Q_t^{\frac{1}{2}} \hat{\xi}_t).$$ We have factored $I + FF^{\top}$ as $LL^{\top}$, so $\gamma^2 I +  G^{\top}(I + FF^{\top})^{-1}G = \gamma^2 I + (L^{-1} G)^{\top} (L^{-1} G)$. Notice that $L^{-1}G$ is strictly causal, since $L^{-1}$ is causal and $G$ is strictly causal. A state-space model for $G$ is $$\eta_{t+1} = A_t \eta_t + B_{w, t} w_t, \hspace{5mm} s_t = Q_t^{\frac{1}{2}} \eta_t.$$ 

\noindent Equating $s$ and $y$, we see that a state-space model for $L^{-1}G$ is  $$\begin{bmatrix} \hat{\xi}_{t+1} \\ \eta_{t+1} \end{bmatrix} = \begin{bmatrix} \tilde{A}_t & K_{p, t}Q_t^{\frac{1}{2}}  \\ 0 & A_t  \end{bmatrix} \begin{bmatrix} \hat{\xi}_{t} \\ \eta_{t} \end{bmatrix} + \begin{bmatrix} 0 \\ B_{w, t}  \end{bmatrix} w_t, $$ $$e_t =  R_{e, t}^{-\frac{1}{2}}Q_t^{\frac{1}{2}}( \eta_t - \hat{\xi}_t), $$
where we defined $\tilde{A}_t = A_t - K_{p, t} Q_t^{\frac{1}{2}}$. Setting $\nu_t = \eta_t - \hat{\xi}_t$ and simplifying, we see that a minimal representation for a state-space model for $L^{-1}G$ is  

\begin{equation} \label{intermediate-ss}
\nu_{t+1} = \tilde{A}_t\nu_t + B_{w, t}w_t, \hspace{5mm} e_t = R_{e, t}^{-\frac{1}{2}} Q_t^{\frac{1}{2}} \nu_t,
\end{equation}
It follows that a state-space model for $(L^{-1}G)^{\top}$ is $$\nu_{t-1} = \tilde{A}_t^{\top}\nu_t + Q_t^{\frac{1}{2}} R_{e, t}^{-\frac{1}{2}}  w_t, \hspace{5mm} e_t =  B_{w, t}^{\top} \nu_t.$$ 

Recall that our original goal was to obtain a factorization  $\gamma^2 I +  G^{\top}(I + FF^{\top})^{-1}G = \Delta^{\top}\Delta$, where $\Delta$ is causal. Let $z = (L^{-1}G)^{\top} a + b$, where $a$ and $b$ are zero-mean random variables such that $\expect[a a^{\top}] = I, \expect[a b^{\top}] = 0, $ and $\expect[b b^{\top}] = \gamma^2 I$. Suppose that we can find an causal matrix $\Delta$ such that $z = \Delta^{\top} f$, where $f$ is a zero-mean random variable such that $\expect[f f^{\top}] = I$. Notice that $\expect[z z^{\top}] = \gamma^2 I + (L^{-1} G)^{\top} (L^{-1} G) =  \gamma^2 I +  G^{\top}(I + FF^{\top})^{-1}G$; on the other hand $\expect[zz^{\top}] = \Delta^{\top}\Delta$ as desired.

A backwards-time state-space model for $z$ is given by $$ \nu_{t-1} = \tilde{A}_t^{\top}\nu_t - Q_t^{\frac{1}{2}} R_{e, t}^{-\frac{1}{2}} a_t, $$ $$ z_{t} = B_{w, t}^{\top} \nu_t + b_t.$$ Using the (backwards time) Kalman filter, we see that a state-space model for $L^{\top}$ is given by $$\hat{\nu}_{t-1} =  \tilde{A}_t^{\top}\hat{\nu}_t + K_{l, t}^b (R_{e, t}^b)^{\frac{1}{2}}f_t, $$ $$ z_t = B_{w, t}^{\top}\hat{\nu}_t  + (R_{e, t}^b)^{\frac{1}{2}}f_t,$$ where we define $K_{l, t}^b = (A_t - K_{p, t}Q_t^{\frac{1}{2}})^{\top}P_t^b B_{w, t} (R_{e, t}^b)^{-1} $  and $R_{e, t}^b = \gamma^2 I + B^{\top}_{w, t} P_t^b B_{w, t}$, and $P_t^b$ is the solution to the backwards Riccati recursion 
$$P_{t-1}^b =  \tilde{A}_t^{\top}P_{t}^b \tilde{A}_t + Q_t^{\frac{1}{2}}(R_{e, t})^{-1}Q_t^{\frac{1}{2}} - K_{l, t}^b R_{e, t}^b (K_{l, t}^b)^{\top}, $$ and $P_T^b = 0$.  It follows that a state-space model for $\Delta$ is given by $$\hat{\nu}_{t+1} =  \tilde{A}_t\hat{\nu}_t + B_{w, t} f_t, $$ $$ z_t = (R_{e, t}^b)^{\frac{1}{2}} (K_{l, t}^b)^{\top}\hat{\nu}_t  +   (R_{e, t}^b)^{\frac{1}{2}}f_t.$$ Therefore a state-space model for $\Delta^{-1}$ is $$ \hat{\nu}_{t+1} =  (\tilde{A}_t - B_{w, t}(K^b_{l, t})^{\top})\hat{\nu}_t + B_{w, t}(R^b_{e, t})^{-\frac{1}{2}}z_t,$$ $$ f_t = -(K^b_{l, t})^{\top}\hat{\nu}_t + (R^b_{e, t})^{-\frac{1}{2}}z_t .$$ 
 
 \noindent Recall that a state-space model for $G$ is given by $$\eta_{t+1} = A_t \eta_t + B_{w, t} w_t, \hspace{5mm} s_t = Q_t^{\frac{1}{2}} \eta_t.$$ Equating $f_t$ and $w_t$, we see that a state-space model for $G\Delta^{-1}$ is $$\hat{\nu}_{t+1} =  (\tilde{A}_t - B_{w, t}(K^b_{l, t})^{\top})\hat{\nu}_t + B_{w, t}(R^b_{e, t})^{-\frac{1}{2}}z_t,$$ $$ \eta_{t+1} = A_t \eta_t - B_{w, t}(K^b_{l, t})^{\top}\hat{\nu}_t + B_{w, t} (R^b_{e, t})^{-\frac{1}{2}}z_t, $$ $$ s_t = Q_t^{\frac{1}{2}} \eta_t.$$
A state-space model for $F$ is given by $$\psi_{t+1} = A_t\psi_t + B_{u, t}u_t, \hspace{5mm} s_t = Q_t^{\frac{1}{2}}\psi_t.$$ Letting $\zeta_t = \eta_t + \psi_t$, we see that a state-space model for the overall system is given by 
 
 \begin{align*}
 \begin{bmatrix} \zeta_{t+1} \\ \hat{\nu}_{t+1} \end{bmatrix} &= \begin{bmatrix} A_t & - B_{w, t}(K^b_{l, t})^{\top} \\ 0 &  \tilde{A}_t - B_{w, t}(K^b_{l, t})^{\top} \end{bmatrix} \begin{bmatrix} \zeta_t \\ \hat{\nu}_t \end{bmatrix}  + \begin{bmatrix} B_{u, t} \\ 0 \end{bmatrix} u_t +  \begin{bmatrix} B_{w, t} (R^b_{e, t})^{-\frac{1}{2}} \\ B_{w, t} (R^b_{e, t})^{-\frac{1}{2}} \end{bmatrix} z_t, \\
 s_t &=  \begin{bmatrix} Q_t^{\frac{1}{2}} & 0 \end{bmatrix} \begin{bmatrix} \zeta_t \\ \hat{\nu}_t \end{bmatrix}.
 \end{align*}
 To derive the regret-optimal controller, we can plug this state-space model into the formula for the optimal $H_{\infty}$ controller given in Theorem \ref{hinf-optimal-thm}. We see that the regret-optimal controller is given by 
$$ u_t = - \hat{H}^{-1}_t \hat{B}_{u, t}^{\top} \hat{P}_{t+1} \left(\hat{A}_t \begin{bmatrix} \zeta_t \\ \hat{\nu}_t \end{bmatrix} + \hat{B}_{w, t} z_t \right), $$
 where we define 
$$\hat{A}_t = \begin{bmatrix} A_t & - B_{w, t}(K^b_{l, t})^{\top} \\ 0 &  \tilde{A}_t - B_{w, t}(K^b_{l, t})^{\top} \end{bmatrix}, $$
$$\hat{B}_{u, t} =  \begin{bmatrix} B_{u, t} \\ 0 \end{bmatrix}, $$
$$\hat{B}_{w, t} =  \begin{bmatrix} B_{w, t} (R^b_{e, t})^{-\frac{1}{2}} \\ B_{w, t} (R^b_{e, t})^{-\frac{1}{2}} \end{bmatrix}, $$
 $$\hat{Q}_t = \begin{bmatrix} Q_t & 0 \\ 0 & 0 \end{bmatrix}, $$
 $$ \hat{H}_t = I + \hat{B}_{u, t}^{\top} \hat{P}_{t+1} \hat{B}_{u, t}, $$ 
 and $\hat{P}_t$ is the solution of the backwards Riccati recursion $$\hat{P}_{t} = \hat{Q}_t + \hat{A}_t^{\top}\hat{P}_{t+1}A_t - \hat{A}_t^{\top} \hat{P}_{t+1} \hat{B}_{u, t} \hat{H}_t^{-1} \hat{B}_{u, t}^{\top} \hat{P}_{t+1} \hat{A}_t $$ and we initialize $\hat{P}_T = 0$.
 We emphasize that the driving disturbance in this system is not $w$, but rather $z = \Delta w$. We already found a state-space model for $\Delta$, so it is easy to see that a state-space model for $z$ is given by 
  $$\delta_{t+1} =  \tilde{A}_t\delta_t + B_{w, t} w_t, $$
 $$ z_t = (R_{e, t}^b)^{\frac{1}{2}} (K_{l, t}^b)^{\top}\delta_t  +   (R_{e,t}^b)^{\frac{1}{2}}w_t, $$
 where we initialize $\delta_0 = 0$.
 The regret bound stated in Theorem \ref{regret-optimal-controller-thm} is immediate: by definition the regret-suboptimal controller at performance level $\gamma$ has dynamic regret at most $\gamma^2 \|w\|_2^2$. To find the regret-optimal controller, we minimize $\gamma$ subject to the constraints described in Theorem \ref{hinf-optimal-thm}.
\end{proof}


\section{Integrating predictions and delay} \label{predictions-delay-sec}
In this section, we extend the results of Sections \ref{regret-optimal-sec} to settings with predictions and delay by using reductions to Theorem \ref{regret-optimal-controller-thm}. While we consider control with predictions and control with delay separately, we emphasize that it easy to extend our results to settings with both predictions and delay by performing one reduction and then the other.

\subsection{Regret-optimal model-predictive control}
Consider a system with dynamics given by the linear evolution equation (\ref{evolution-eq}). In model-predictive control we assume that the controller can predict future disturbances over a horizon of length $h$; in other words, we assume that at time $t$ the controller knows the disturbances $w_{t}, \ldots w_{t+h -1}$.  Define the augmented state $$\xi_t = \begin{bmatrix} x_t \\ w_{t} \\ w_{t+1} \\ \vdots \\ w_{t + h -2} \\ w_{t+h -1} \end{bmatrix}. $$ We can think of $\xi_t$ as representing the state $x_t$ along with a transcript of the next $h$ predicted disturbances $w_{t}, \ldots, u_{w+h-1}$. Notice that $\xi$ has dynamics given by the linear evolution equation 

\begin{equation} \label{transformed-pred-eq}
\xi_{t+1} = \hat{A}_t \xi_t + \hat{B}_{u, t}u_t + \hat{B}_{w, t}w_t',
\end{equation}
where we define $w_t' = w_{t+h}$ and
\begin{align*}
\hat{A}_t = \begin{bmatrix} A_t & B_{w, t}  & 0 & \ldots & 0 & 0 \\ 0 & 0 & I & \ldots & 0 & 0 \\ 0  & 0 &  &  &  & 0 \\ \vdots & \vdots  &  & \ddots  & \ddots &  \\ \vdots & \vdots  & & &  & I \\ 0 & 0 & 0 & \ldots & 0 & 0 \end{bmatrix}, \hspace{5mm}  \hat{B}_{u, t} = \begin{bmatrix} B_{u, t} \\ 0 \\ 0 \\  \vdots \\ 0 \\ 0 \end{bmatrix} u_t, \hspace{5mm} \hat{B}_{w, t} = \begin{bmatrix} 0 \\ 0 \\ 0 \\ \vdots \\ 0 \\ I \end{bmatrix}.
\end{align*} 
We can rewrite the LQR cost  in the original system (\ref{evolution-eq}) in terms of $\xi$: 
\begin{equation} \label{new-lqr-cost-predictions}
x_T^{\top}Q_T x_t + \sum_{t=0}^{T-1} \left( x_t^{\top}Q_t x_t + u_t^{\top}R_t u_t \right) = \xi_T^{\top}\hat{Q}_T \xi_t + \sum_{t=0}^{T-1} \left( \xi_t^{\top}\hat{Q}_t \xi_t + u_t^{\top}R_t u_t \right),
\end{equation}
where we define the block-diagonal matrix $$\hat{Q}_t = \begin{bmatrix} Q & 0 & \ldots \\ 0 & 0 \\ \vdots & & \ddots \end{bmatrix}.$$ In light of (\ref{new-lqr-cost-predictions}), we see that any sequence of control actions $u = (u_0, \ldots, u_{T-1})$ generates an identical cost in the original system (\ref{delay-eq}) and in the new system (\ref{transformed-delay-eq}). Notice that a causal control policy in the new system (i.e. one whose control actions at time $t$ depend only on $w_0', \ldots w_t')$ is a policy with a lookahead of length $h$ in the original system. We have proven: 
 
\begin{theorem} \label{regret-optimal-mpc-theorem}
The regret-optimal model-predictive controller with lookahead of length $h$ in the dynamical system (\ref{evolution-eq}) is the regret-optimal controller described in Theorem \ref{regret-optimal-controller-thm} applied to the system (\ref{transformed-pred-eq}). 
 \end{theorem}

\subsection{Regret-optimal control with delay}
Consider the linear evolution equation 

\begin{equation} \label{delay-eq}
x_{t+1} = A_t x_t + B_{u, t-d}u_{t-d} + B_{w, t} w_t.
\end{equation}
In this system, control actions affect the state only after a delay of length $d$; in other words, at time $t$, the state $x_t$ is a function of $w_0, w_1, \ldots w_{t-1}$ and $u_0, u_1, \ldots u_{t-d-1}$. For $t = 0, \ldots T-1$, we define the augmented state $$\xi_t = \begin{bmatrix} x_t \\ u_{t-1} \\ u_{t-2} \\ \vdots \\ u_{t-d + 1} \\ u_{t-d} \end{bmatrix}. $$ We can think of each $\xi_t$ as representing the actual state $x_t$ along with a transcript of the previous $d$ control actions. Notice that $\xi$ has dynamics given by the linear evolution equation 
\begin{equation} \label{transformed-delay-eq}
\xi_{t+1} = \hat{A}_t \xi_t + \hat{B}_{u, t} u_t + \hat{B}_{w, t}w_t,
\end{equation}
where we define 
\begin{align*}
\hat{A}_t =  \begin{bmatrix} A_t & 0  & 0 & \ldots & 0 & B_{u, t - d} \\ 0 & 0 & 0 & \ldots & 0 & 0 \\ 0  & I &  0 & \ldots & 0 & 0 \\ \vdots &  &  \ddots & &  & \vdots \\ 0 &  & & \ddots &  & \vdots \\ 0 & 0 & 0 & \ldots & I & 0 \end{bmatrix}, \hspace{5mm} \hat{B}_{u, t} =  \begin{bmatrix} 0 \\ I \\ 0 \\  \vdots \\ 0 \\ 0 \end{bmatrix}, \hspace{5mm} \hat{B}_{w,t} =  \begin{bmatrix} B_{w, t} \\ 0 \\ 0 \\ \vdots \\ 0 \\ 0 \end{bmatrix}.
\end{align*}
We can rewrite the LQR cost in the original system (\ref{delay-eq}) in terms of $\xi$: 
\begin{equation} \label{new-lqr-cost-delay}
\sum_{t=0}^{T-1} \left( x_t^{\top}Q_t x_t + u_t^{\top}R_t u_t \right) = \xi_T^{\top}\hat{Q}_T \xi_t + \sum_{t=0}^{T-1} \left( \xi_t^{\top}\hat{Q}_t \xi_t + u_t^{\top}R_t u_t \right),
\end{equation}
where we define the block-diagonal matrix $$\hat{Q}_t = \begin{bmatrix} Q & 0 & \ldots \\ 0 & 0 \\ \vdots & & \ddots \end{bmatrix}.$$ In light of (\ref{new-lqr-cost-delay}), we see that any sequence of control actions $u = (u_0, \ldots, u_{T-1})$ generates an identical cost in the original system (\ref{delay-eq}) and in the new system (\ref{transformed-delay-eq}). Also notice that the new system is of the form (\ref{evolution-eq}), i.e. it has no delay. A causal controller in the new system is a causal controller in the original system. We have proven: 
 
\begin{theorem} \label{regret-optimal-delay-theorem}
The regret-optimal controller in the dynamical system with delay (\ref{delay-eq}) is the regret-optimal controller described in Theorem \ref{regret-optimal-controller-thm} applied to the system (\ref{transformed-delay-eq}). 
\end{theorem}

\section{Numerical experiments} \label{numerical-experiments-sec}

\begin{figure} 
    \begin{center}
    \includegraphics[scale=0.28]{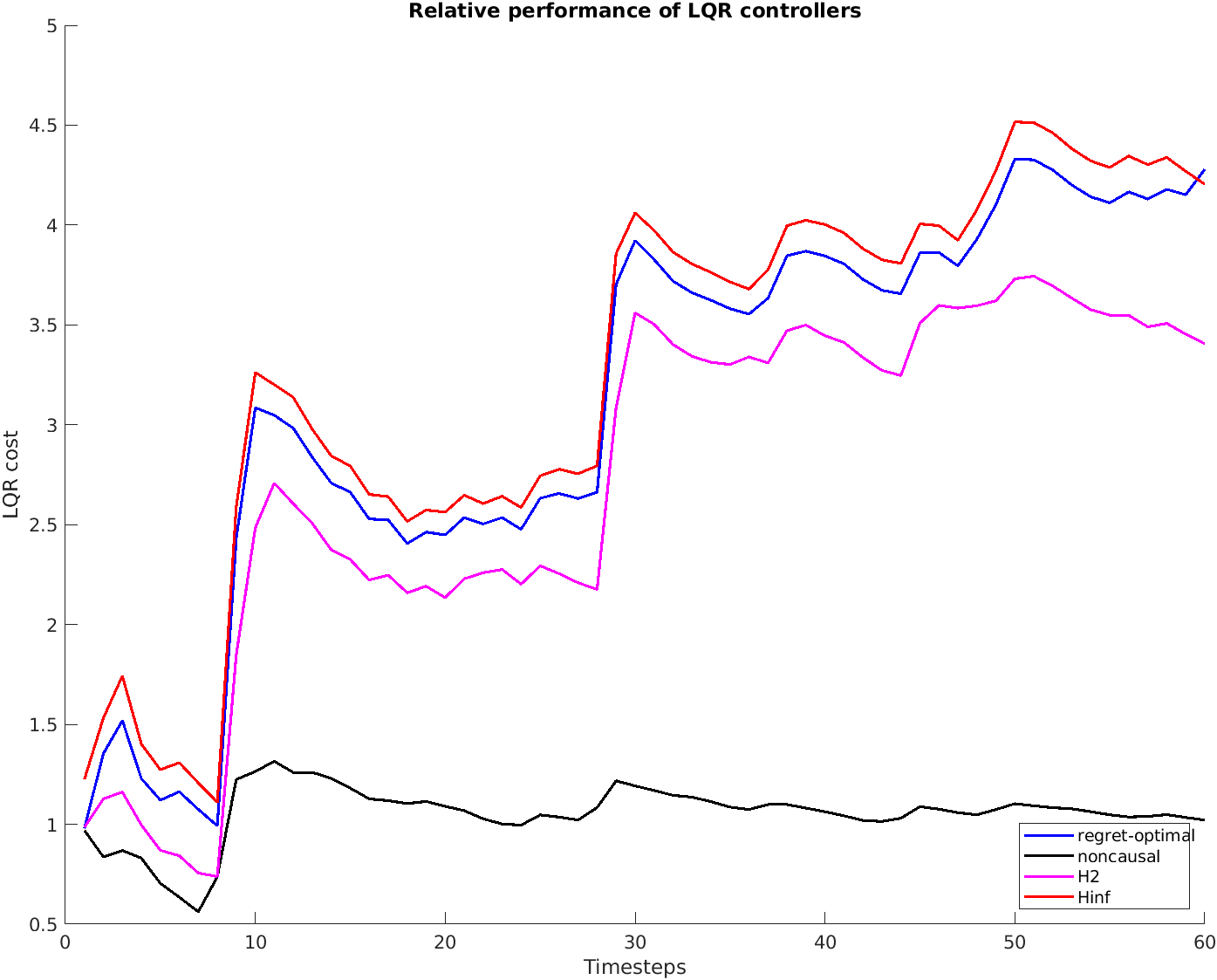}
    \hspace{2mm}
    \includegraphics[scale=0.28]{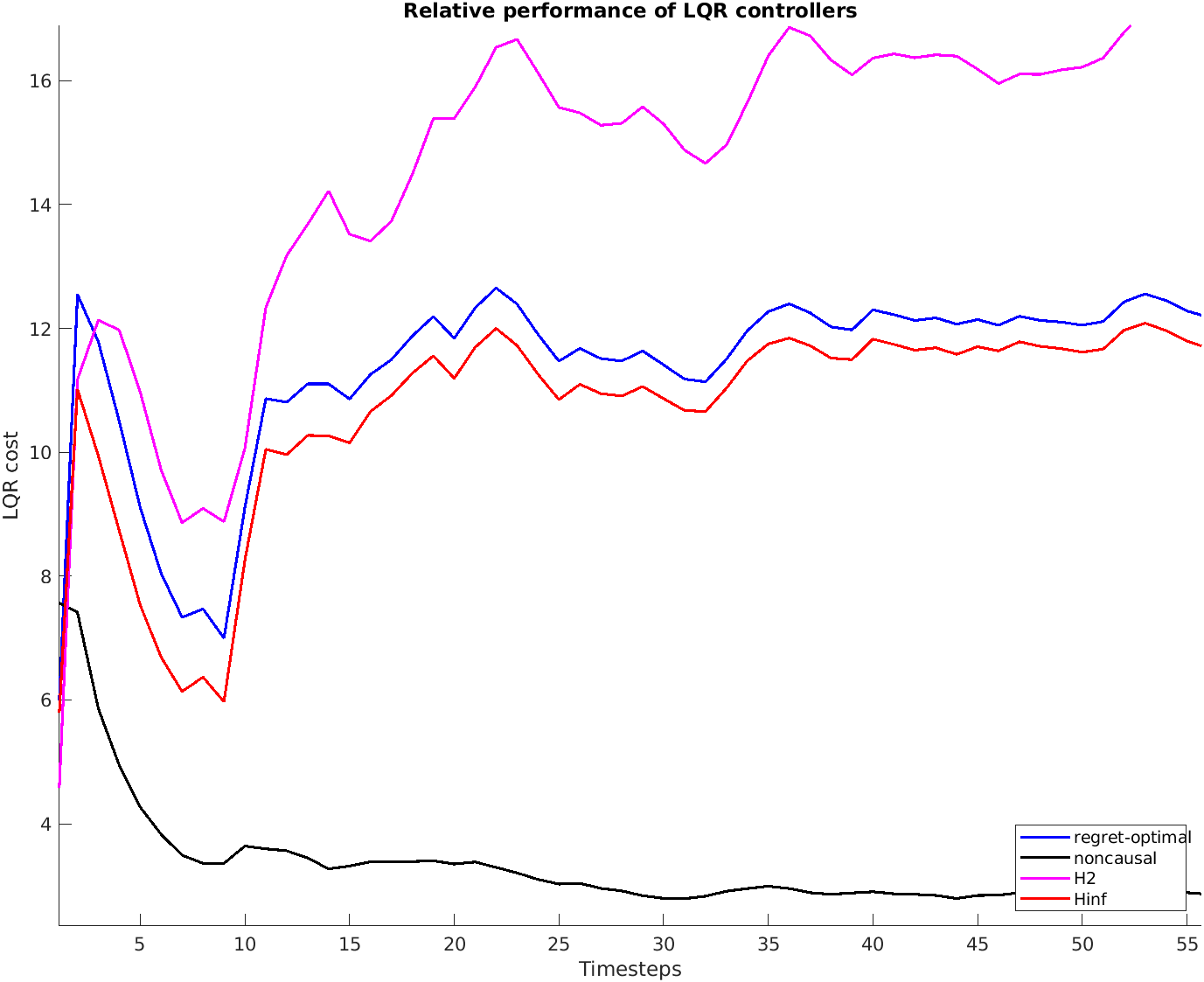}
    \end{center}
    \caption{We plot the time-averaged costs of the LQR controllers with noisy inverted pendulum dynamics. The first plot shows that when each component of the noise is drawn independently from $\mathcal{N}(0, 1)$ the regret-optimal controller's performance lies between that of the $H_2$-optimal controller and the $H_{\infty}$-optimal controller. In the second plot, the noise components alternate between being drawn from $\mathcal{N}(1, 1)$ and  $\mathcal{N}(-1, 1)$ every 15 timesteps. We see that the regret-optimal controller significantly outperforms the $H_2$-optimal controller and closely tracks the performance of the $H_{\infty}$-optimal controller. }
    \label{inv-pend-figure}
\end{figure}

We refer the reader to Examples 2.2 and 4.4 in \cite{aastrom2021feedback} for background on the physics of the inverted pendulum; for our purposes it suffices to know that the continuous-time closed-loop dynamics of the inverted pendulum is given by the nonlinear differential equation $$\frac{dx}{dt} = \begin{bmatrix} x_2 \\ \sin{x_1} - c x_2  + u \cos{x_1} \end{bmatrix}, $$ where $x = (x_1, x_2)$ is the state, $u$ is a scalar control input, and $c$ is a physical parameter of the system. Using the linearizations $\sin{x} \approx x, \cos{x} \approx 1$ for small $x$, and $x_{t+\Delta} \approx x_{t} + \frac{dx}{dt}\Delta$ for small $\Delta$ and adding a disturbance term, we can capture these dynamics by the state-space model $$x_{t+1} = \begin{bmatrix} 1 & 1 \\ 1 & 1-c \end{bmatrix}x_t +  \begin{bmatrix} 0 \\ 1 \end{bmatrix}u_t  + w_t .$$ In our experiments we set $ c = 0.1$, $Q = I, R = 1$ and initialize $x_0 = 0$. Note that $w_t \in \mathbb{R}^2$; in our experiments we set each of the two components individually. In our first experiment, we consider a stationary stochastic environment and draw each component independently from $\mathcal{N}(0, 1)$ in each timestep. In our second experiment, we consider a more adversarial setting and alternate between drawing from $\mathcal{N}(1, 1)$ and  $\mathcal{N}(-1, 1)$ every 15 timesteps. Figure \ref{inv-pend-figure} shows that the regret-optimal controller interpolates between the performance of the $H_2$-optimal and $H_{\infty}$-optimal controllers.

\end{document}